\documentclass[11pt,a4paper]{article}
\usepackage[hyperref]{emnlp2020}
\usepackage{times}
\usepackage{latexsym}

\usepackage{microtype}

\usepackage{comment}
\usepackage{multirow}
\usepackage{bm}

\usepackage{amsmath,amssymb}

\usepackage{graphicx}

\usepackage{booktabs}
\usepackage{multirow}
\usepackage{url}
\usepackage{todonotes}
\usepackage{enumitem}

\aclfinalcopy %

\title{Examining the Ordering of Rhetorical Strategies in Persuasive Requests}

\author{Omar Shaikh, Jiaao Chen, Jon Saad-Falcon, Duen Horng (Polo) Chau, Diyi Yang \\
  Georgia Institute of Technology \\
  \texttt{\{oshaikh, jchen896, jonsaadfalcon, polo, dyang888\}@gatech.edu}  \\}

\date{}
\usepackage{pifont}%

\begin{document}
\maketitle
\begin{abstract}
Interpreting how persuasive language influences audiences has implications across many domains like advertising, argumentation, and propaganda. Persuasion relies on more than a message's content. Arranging the order of the message itself (i.e., ordering specific rhetorical strategies) also plays an important role.  To examine how strategy orderings contribute to persuasiveness, we first utilize a Variational Autoencoder model to disentangle content and rhetorical strategies in textual requests from a large-scale loan request corpus. We then visualize interplay between content and strategy through an attentional LSTM that predicts the success of textual requests. We find that specific (orderings of) strategies interact uniquely with a request's content to impact success rate, and thus the persuasiveness of a request.

\end{abstract}

\section{Introduction}
Persuasion has been shown as a powerful tool for catalyzing beneficial social and political changes \citep{hovland1953communication} or enforcing propaganda as a tool of warfare \citep{lyneyve2000}. Modeling persuasiveness of text has received much recent attention in the language community \citep{althoff2014ask, Tan_2016, habernal-gurevych-2017-argumentation, yang-etal-2019-lets, 10.1145/3359265}.
Numerous qualitative studies have been conducted to understand persuasion, from explorations of rhetoric in presidential campaigns \citep{bartels, popkin1994reasoning} to the impact of a communicator's likability on persuasiveness~\citep{chaiken1980heuristic}. 
Studies of persuasion and argumentation that have analyzed textual level features (e.g., n-grams, independent rhetorical strategies) to gauge efficacy have also garnered recent attention~\citep{althoff2014ask, habernal-gurevych-2017-argumentation, habernal-gurevych-2016-argument, habernal-gurevych-2016-makes,yang2017persuading, yang-etal-2019-lets}. Of particular interest is \citet{morio-etal-2019-revealing}, which identified sentence placements for individual rhetorical strategies in a request. Other research analyzed how different persuasive strategies are more effective on specific stances and personal backgrounds~\citep{durmus-cardie-2018-exploring, durmus-cardie-2019-corpus}. 

\begin{table}[]
\renewcommand{\arraystretch}{1.1}
\setlength{\tabcolsep}{2pt}
\small
\centering
\begin{tabular}{@{}ll@{}}
\toprule
Strategy & Definition \\ %
\midrule
\begin{tabular}[t]{@{}l@{}}\textbf{Co}ncreteness
(39\%) \end{tabular}& \begin{tabular}[t]{@{}l@{}} Use concrete details in request\\ \emph{``I need \$250 to purchase fishing rods''}\\  \end{tabular} \\ %
\begin{tabular}[t]{@{}l@{}}\textbf{Re}ciprocity (18\%) \end{tabular}& \begin{tabular}[t]{@{}l@{}} Assure user will repay giver \\ \emph{``I will pay 5\% interest to you''} \end{tabular}\\ %
\begin{tabular}[t]{@{}l@{}}\textbf{Im}pact (12\%)\end{tabular} & \begin{tabular}[t]{@{}l@{}} Highlight the impact of a request \\\emph{``This loan will help teach students''}\end{tabular} \\ %
\begin{tabular}[t]{@{}l@{}}\textbf{Cr}edibility  (8\%) \end{tabular}& \begin{tabular}[t]{@{}l@{}} Use credentials to establish trust \\ \emph{``I have repaid all of my prior loans''} \end{tabular} \\ %
\begin{tabular}[t]{@{}l@{}}\textbf{Po}liteness (16\%) \end{tabular}& \begin{tabular}[t]{@{}l@{}}Use polite language \\ \emph{``Highly appreciated.''} \end{tabular}\\ %
Other (7\%) & None of the above \\ %
\bottomrule
\end{tabular}
\caption{\small Sentence level persuasion strategies, and their data distributions (\%). Strategy abbreviations are \textbf{bolded}.}
\label{sentlevelstrats}
\end{table}

\begin{figure*}
\centering
  \includegraphics[width=1\linewidth]{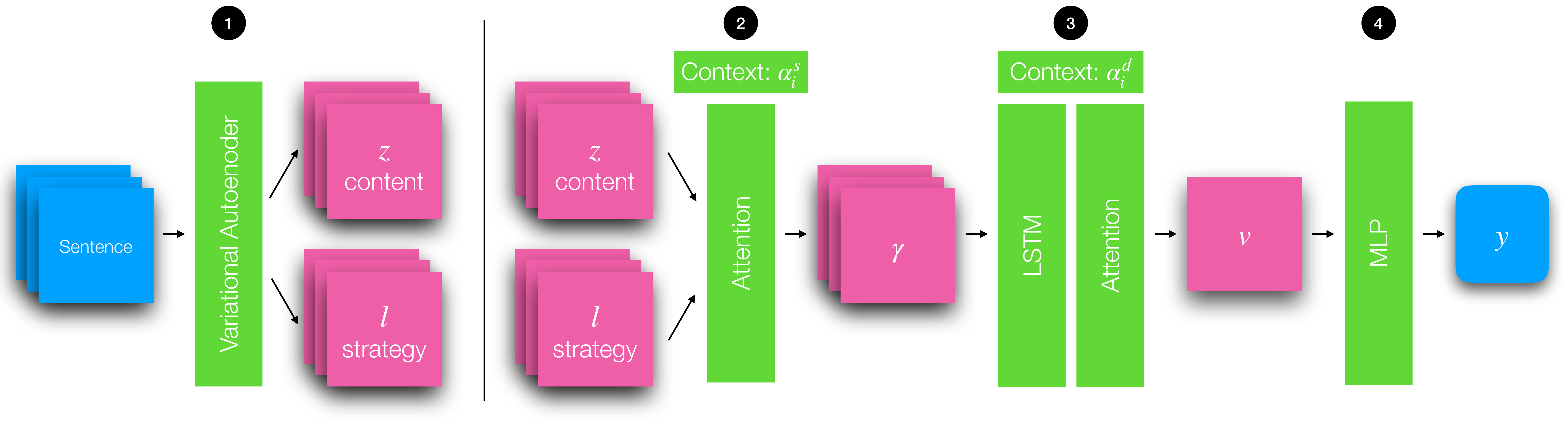}
  \caption{\small Our modeling setup, detailed in section \ref{method-section}, consists of 4 steps. Step 1 deconstructs sentences into latent content and strategy vectors, using a semi-supervised VAE; Step 2 combines content and strategy vectors at the sentence level, using sentence level attention; Step 3 uses an LSTM to model our sentences in a request, then combines sentences using request level attention. Finally, Step 4 predicts our binary persuasiveness label using a multilayer perceptron. 
  }
\end{figure*}

However, prior work has mainly focused on identifying overall persuasiveness of textual content or analyzing components of persuasion affecting a request. These works largely ignore \textbf{ordering} of specific strategies, a key canon of rhetoric that has a large impact on persuasion effectiveness \citep{borchers2018rhetorical, cicero1862cicero}.
In the context of online communities, identifying where/how effective orderings occur may highlight qualities of persuasive requests and help users improve their rhetorical appeal. Furthermore, highlighting ineffective orderings may help users avoid pitfalls when framing their posts.

To fill this gap, we propose to investigate particular orderings of persuasive strategies that affect a request's persuasiveness and identify situations where these orderings are optimal. %
Specifically, we take a closer look at strategies (Table \ref{sentlevelstrats}) and their orderings in requests from the subreddit/online lending community \textbf{r/Borrow} \footnote{\small{\url{https://www.reddit.com/r/borrow/}}}; and utilize them to examine research questions like: When should requesters follow strategy orderings (e.g., ending loan requests with politeness) that rely on social norms? Should requesters worry less about orderings and more about content? Altogether, this work examines \textbf{orderings}, an overlooked rhetorical canon, and how they interact with a request's persuasiveness in an online lending domain.
Our contributions include: 
\begin{enumerate}
    \item Identifying specific strategy orderings that correlate with requests' persuasiveness.
    \item Highlighting the interplay between content and strategy with respect to the persuasiveness of a request.
    \item Perturbing underperforming strategy orderings to help improve persuasiveness of requests via a set of introduced edit operations.
\end{enumerate}
Code for our analyses can be found at \url{https://github.com/GT-SALT/Persuasive-Orderings}.

\section{Method}
\label{method-section}

\subsection{Dataset} 
Our Borrow dataset consists of 49,855 different loan requests in English, scraped from the r/Borrow subreddit. r/Borrow is a community which financially assists users with small short-term loans to larger long-term ones. Every request has a binary label indicating if a loan is successful or not. Request success rate, on average, is 48.5\%. We randomly sampled a subset (5\%) from the whole corpus to annotate their sentence-level labels indicating persuasive strategies; labels were adapted from \citet{yang-etal-2019-lets} (and defined in Table \ref{sentlevelstrats}). 
We recruited four research assistants to label persuasion strategies for each sentence. Definitions and examples of different persuasion strategies were provided, together with a training session where we asked annotators to annotate a number of example sentences and walked them through any disagreed annotations. To assess the reliability of the annotated labels, we then asked them to annotate a small subset of 100 requests from our corpus, with a Cohen' Kappa of .623, indicating moderate annotation agreement \citep{mchugh2012interrater}.
Annotators then annotated the rest of corpus by themselves independently. 
In total, we gathered 900 requests with sentence-level labels and 48,155 requests without sentence-level labels as our training set, 400 requests with sentence-level labels as the validation set and 400 requests with sentence-level labels as the test set.

\subsection{Modeling}
Persuasive sentences are combinations of content (what to include in persuasive text) and strategy (how to be persuasive). To explore interplay between content and strategy orderings inside requests, we followed \citet{kingma2013auto} and \citet{yang2017improved}, utilizing a semi-supervised Variational Autoencoder (VAE) trained on both labeled and unlabeled sentences to disentangle sentences into strategy and content representations. Specifically, for every input sentence ${x}$, we assumed the graphical model $p({x}, {z}, {l}) = p({x}|{z}, {l})p({z})p({l})$, where ${z}$ is a latent ``content'' variable and ${l}$ is the persuasive strategy label.
The semi-supervised VAE fits an inference network $q({z}|{x}, {l})$ to infer latent variable ${z}$, a generative network $p({x}|{l},{z})$ to reconstruct input sentence ${s}$, and a discriminative network $q({l}|{x})$ to predict persuasive strategy ${l}$, while optimizing an evidence lower bound (ELBO) similar that of general VAE. 
We report a Macro F-1 score of $0.75$ on the test set for \emph{sentence-level classification}, suggesting reasonable performance compared to an LSTM baseline with a Macro F-1 of $0.74$ \citep{yang-etal-2019-lets}. %
Then, for each request $M = \{x_0, x_1, ..., x_L\}$ consisting of $L$ sentences that a user posted to receive a loan, we utilized our trained semi-VAE to represent each sentence $x_i$ in $M$ with content and strategy variables to form  $M’ = \{(z_0, l_0), (z_1, l_1) …(z_L, l_L)\}$.

With the intent of interpreting importance between strategy orderings and content, we built an attentional LSTM trained to predict success of a request.  For each disentangled sentence $(z_i, l_i)$ in our requests, we first applied attention on $z_i$ and $l_i$ at the \textbf{sentence level}, dynamically combining them into sentence representation $\gamma_{i}$:
\begin{align*}
    u_{z_{i}} &=\tanh \left(W_{z} z_{i}+b\right) \\
    u_{y_{i}} &=\tanh \left(W_{l} y_{i}+b\right) \\
\alpha^s_{i} &= \frac{\langle\exp \left(u_{z_{i}}^{\top} u_{q}\right), \exp \left(u_{l_{i}}^{\top} u_{q}\right)\rangle}{\exp \left(u_{z_{i}}^{\top} u_{q}\right) + \exp \left(u_{l_{i}}^{\top} u_{q}\right)}\\
\gamma_{i} &= \left(\alpha^s_{i, 0} \cdot z_{i}\right) \oplus \left(\alpha^s_{i, 1} \cdot l_{i}\right)
\end{align*}
where $u$ are randomly initialized context vectors that were jointly learned with weights $W$.
We computed the request representation $v$ through an LSTM that encoded sentence representations $\gamma_{i}$ for each request, and a \textbf{request level} attention that aggregated information from different sentences. Overall persuasiveness is predicted as:
\begin{align*}
u_{i} &=\tanh \left(W_{s} h_{i}+b_{s}\right) \\
\alpha^d_{i} &= \frac{\exp \left(u_{i}^{\top} u_{s}\right)}{\sum_{k} \exp \left(u_{k}^{\top} u_{s}\right)} \\
v &=\sum_{i} \alpha^d_{i} h_{i}\ \  \text{and} \   
y = \text{MLP}(v)
\end{align*}

The training objective is regular cross entropy loss. Macro-averaged performances for \emph{request-level classification} on several baseline classifiers are shown in Table \ref{baselineperf}. 
Our attentional model (VAE + LSTM) achieves comparable performance to BERT, while providing additional benefit of disentangling content and strategy. This helps yield relative measures of importance for content and strategies. 
\section{Interplay of Ordering and Content}

\begin{table}[]
\centering
\small %
\begin{tabular}{@{}lrrr@{}}  %
\toprule
Model &  F-1 & Precision & Recall \\ 
\midrule
Naive Bayes & .60 & .60 & .60  \\
\begin{tabular}[t]{@{}l@{}}BERT\end{tabular} & .65 & .64 & .65 \\
\textbf{VAE + LSTM} & .61 & .61 & .61\\
\bottomrule
\end{tabular}
\caption{Request label performance on test set.}
\label{baselineperf}
\end{table}

\begin{table*}[]
\centering
\small
\resizebox{\textwidth}{!}{\
\begin{tabular}[t]{@{}llrr@{}}
\toprule
Strategy & Example & \begin{tabular}[t]{@{}r@{}}Strategy \\ Attention\end{tabular} & \begin{tabular}[t]{@{}r@{}}Success \\ Rate\end{tabular} \\ %
\midrule
 (Po, Po, EOS) & \begin{tabular}[t]{@{}l@{}} ... Your help is \textbf{deeply appreciated!} \textbf{Thank you} for reading and considering my request!\end{tabular} & .00 & .82 \\ \cmidrule{2-2}
(Re, Po, EOS) & \begin{tabular}[t]{@{}l@{}} ... I would greatly appreciate the help, will \textbf{pay back latest on 10/31/18 with interest} determined \\ by the lender. \textbf{Thank you in advance!}\end{tabular} & .01 & .76 \\ 
\cmidrule{2-2}
 (Co, Po, EOS) & \begin{tabular}[t]{@{}l@{}} ... I’m trying to budget myself to where I can borrow \textbf{less this pay period} and get myself to where \\ I won’t need to borrow anymore. Until then I would really \textbf{appreciate the help!} \end{tabular} & .02 & .71  \\ 
 \cmidrule{0-3}
(Co, Im, Co) & \begin{tabular}[t]{@{}l@{}} ... I am switching jobs and need to do \textbf{2 weeks off site training.} The store I am training at is several \\ hours away so I need gas money and money \textbf{for food} while I am out of town. Everything I spend \\ will be \textbf{reimbursed at the end of training}... \end{tabular} & .09 &.34 \\ %
\cmidrule{2-2}
 (Co, Co, Co) & \begin{tabular}[t]{@{}l@{}} ... Was supposed to be on a flight back to TX \textbf{earlier today.} It got canceled \textbf{due to weather and delays}\\ (can provide proof). I’m now stuck in Colorado for \textbf{another day} before I can get back to work ...\end{tabular} & .11 & .31  \\ %
 \cmidrule{2-2}
 (Co, Co, Im) & \begin{tabular}[t]{@{}l@{}} ... Hello, \textbf{first time} borrow. Requesting \textbf{150\$ because} I just spent my savings on fixing my \\ transmission on my car as well as paying for classes and recently I just paid to pay a citation fee I got. \\ At this moment I really need this just to \textbf{help with bills \& food} while classes are starting next week... \end{tabular} & .11 & .27 \\ %
 \bottomrule
\end{tabular}}
\caption{\small Top 3 followed by Bottom 3 strategy triplets with average strategy attentions.
Strategies are: \textbf{Co}ncreteness, \textbf{Re}ciprocity, \textbf{Im}pact, \textbf{Cr}edibility, and \textbf{Po}liteness. \textbf{SOS} indicates start of request; \textbf{EOS} indicates the end of a request.}
\label{borrow_strats}
\end{table*}

To examine how different strategy orderings contribute to overall persuasiveness of requests, we identified relationships between strategy orderings and success rate by analyzing learned attention weights between strategy orderings and content in our model. Motivated by the ``Rule of Three'' prevalent in persuasive writing \citep{clark_2016}, we utilized triplets as our strategy unit of analysis. The most important strategy triplet in each request was considered to be its \textbf{``persuasion strategy triplet."} 

Pinpointing strategy triplets involved finding the most important consecutive three sentences $((z_{m-1}, l_{m-1}), (z_m, l_m), (z_{m+1}, l_{m+1}))$ in one request based on highest \textbf{request-level} ($\alpha^{d}$) attention weight associated with a sentence. The strategies $(l_{m - 1}, l_m, l_{m + 1})$ associated with these sentences were defined as the aforementioned strategy triples. We noted that the cumulative request-level ($\alpha^{d}$) attention placed on strategy triplets had $\mu=.98$ and $\sigma=.07$, indicating that a single triplet carried most responsibility for persuasiveness of requests. For our analysis, we also defined \textbf{success rate} of a strategy triplet as the average success rate of the requests it belongs to, irrespective of how important it is to a request (ignoring $\alpha^{d}$). To control for infrequent triplets, we defined rare strategies as consisting of less than $0.5$\% of our dataset. We filtered these rare strategies, along with triplets containing the undefined ``Other'' strategy. %
Finally, we averaged \textbf{sentence-level} attention weights $\alpha^s$ on each strategy representation in a triplet to represent the importance of an ordering pattern compared to content. 
Figure \ref{strategyAttnRelationship} plots \textbf{sentence-level} attention weights for each strategy triplet and its corresponding success rate.

We made three discoveries. 
(1) Success rate and triplet attention were \textit{strongly negatively correlated} ($R=-.90$, $p<.0001$). Therefore, the model paying larger attention to strategy triplets may communicate a request's lack of persuasiveness.
(2) Attention from around strategy (SOS, Im, Re) onward decreased substantially, suggesting that \emph{content (complementary to strategy attention) played an increasingly larger role in determining the persuasiveness for strategies triplets above average success rate.}
(3) Under-performing strategies \emph{actively decreased} request persuasiveness, \emph{sabotaging its success}; an under-performing strategy ordering pattern with any content often resulted in reduced persuasiveness. On the contrary, simply having an over-performing strategy with respect to the average success rate does not appear to affect a request due to reduced attention.

We also manually examined around 300 examples,
with representative ones shown in Table~\ref{borrow_strats}. Generally, over-performing triplets had little effect on the success rates due to reduced strategy attention. However, under-performing triplets were relatively highly attended to. Below, we explain two general situations that highlight an over-performing and under-performing strategy pattern from a social science perspective:

\begin{figure}
\centering
  \includegraphics[width=0.95\linewidth]{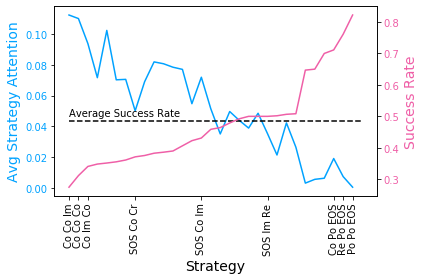}
  \caption{\small Strategy attention triplet vs success rate. The X-axis represents selected strategy triplets, sorted by success rate. A complete list of triplets can be found in Appendix.
  }
  \label{strategyAttnRelationship}
\end{figure}

\section{Common Persuasive Patterns}

\subsection{``Please sir, I want some more.''} A common pattern among the top 5 strategy triplets is the use of politeness. Oftentimes, the politeness triplet appears at the end of the sentence and is usually paired with some form of reciprocity. From Figure \ref{strategyAttnRelationship}, we observed that the best strategy---(Po, Po, EOS)---is a triplet with higher success rates than the average. From a social science perspective, ending a request politely engenders a sense of liking and creates connections between the audience and requester, consistent with prior work showing that politeness is a social norm associated with ending a conversation \citep{schegloff1973opening}. An example is shown in the first row in Table~\ref{borrow_strats}. 
However, this strategy alone does not result in a persuasive request as its associated strategy attention is relatively low. Users who end requests politely \textit{may be likely to put effort into content,} aligning with our success rate observations. Adding to \citet{althoff2014ask}, we observed that users who exercise social ``strategy'' norms by closing conversations politely are shifting importance of a request from strategy to content. Thus, content must still be optimal for a request to be persuasive.

\noindent
\subsection{``It's My Money \& I Need It Now.''} On the contrary, if a triplet consists mostly of concreteness, it performs far below average. For instance, triplets like (Co, Co, Co) often came up in examples that were demanding as shown in Table~\ref{borrow_strats}. From a social science perspective, emotional appeal in arguments is key to framing aspects of a request and helps soften attention placed on facts \citep{walton_1992, macagno_walton_2014, oraby-etal-2015-thats}. In the context of our dataset---a lending platform where concreteness consists mostly of demands---a lack of emotive argumentation may cause an audience to focus on demands themselves, resulting in concrete and emotionless requests.  %
\section{Improving Request Persuasiveness}
\subsection{Editing Operations}
Based on the effectiveness of different persuasion patterns we discovered, this section examines improving underperforming persuasion requests by editing the persuasion strategy patterns. Here we define three editing operations: (1) \textbf{Insert} \citep{ebrahimi-etal-2018-hotflip, wallace2019universal} overperforming triplets into the end of less persuasive requests containing underperforming triplets. (2) \textbf{Delete} \citep{ebrahimi-etal-2018-hotflip} the underperforming triplets in less persuasive requests. (3) \textbf{Swap} \citep{ebrahimi-etal-2018-hotflip} underperforming triplets by first deleting the underperforming triplets and then appending overperforming triplets to the request. We also noticed that overperforming triplets were general conversation closers; their insertion at the end of requests would not alter the intent of a message. 
We performed editing operations to the least persuasive requests (974 examples) containing the bottom 3 underperforming triplets---(Co, Im, Co), (Co, Co, Co), (Co, Co, Im)---using a randomly sampled triplet from the top 3 overperforming triplets (775 examples)---(Po, Po, EOS), (Re, Po, EOS), (Co, Po, EOS).  Table~\ref{pertrubResults} summarizes the results.
\subsection{Editing Results}
For \textbf{Insertion}, the underperforming requests did not improve by simply inserting a good ending triplet, partially because the underperforming request already consisted of a \textsl{sabotaging strategy}; %
furthermore, audiences are likely to generalize impressions from an underperforming strategy to the entire request \citep{ambady1992thin}. %

\textbf{Deletion} of the poor strategy triplets boosted the persuasiveness of a request by mitigating the sabotaging effects of a non-persuasive strategy; however, since the content of the remaining request is mainly unedited, these request still have lower success rates than naturally occurring triplets. 

\textbf{Swapping} the underperforming triplets with effective strategy triplets generated similar persuasiveness to deletion, suggesting again that the presence of an overperforming strategy triplet \emph{does not} improve the persuasiveness of a request (unlike the sabotaging nature of an underperforming triplet); instead, it signals that a given request naturally contains good content since users who put effort into following social norms will likely work hard on the content. This may explain why requests that naturally contain overperforming triplets have higher success rates than our edited examples. Simply swapping strategies does not improve the content, and thus the persuasiveness, to a similar extent.
\begin{table}[]
\centering
\small
\begin{tabular}{@{}lrrr@{}}
\toprule
 & Insert & Delete & Swap\\ %
\midrule
Predicted Average Success Rate & $.11$ & $.43$ & $.42$\\
$\Delta$ from Original Request & $+.00$ & $+.32$ & $+.31$\\
\bottomrule
\end{tabular}
\caption{\small Predicted success rate and the average attention after our editing operations ($\mu$ over 30 runs).}
\label{pertrubResults}
\end{table}
\section{Conclusion \& Future Work}
In this work, we highlight important strategy orderings for request persuasiveness, and surface complex relationships between content and strategy at different request success rates. Finally, we notice improvements in persuasiveness by editing underperforming strategies. For future work,
we plan to explore different techniques for explainability other than attention and compare effective strategies beyond the triplet level across different datasets. We also aim to look at the presence of different strategies across multi-modal settings; does introducing a new modal affect how effective/ineffective strategies are expressed? Furthermore, we plan to identify and compare different strategies across domains, as our work is limited to lending platforms---we expect that different domains would highlight different strategies.%

\section*{Acknowledgements}
We would like to thank the anonymous reviewers
for their helpful comments. 
We thank the members of Georgia Tech's SALT Lab and Dr. Lelia Glass for their feedback on this work. %
DY is supported in part by a grant from Google.
This work is supported in part by NSF grants IIS-1563816, CNS-1704701.
\bibliography{emnlp2020}

\begin{thebibliography}{34}
\expandafter\ifx\csname natexlab\endcsname\relax\def\natexlab#1{#1}\fi

\bibitem[{Althoff et~al.(2014)Althoff, Danescu-Niculescu-Mizil, and
  Jurafsky}]{althoff2014ask}
Tim Althoff, Cristian Danescu-Niculescu-Mizil, and Dan Jurafsky. 2014.
\newblock How to ask for a favor: A case study on the success of altruistic
  requests.
\newblock In \emph{Eighth International AAAI Conference on Weblogs and Social
  Media}.

\bibitem[{Ambady and Rosenthal(1992)}]{ambady1992thin}
Nalini Ambady and Robert Rosenthal. 1992.
\newblock Thin slices of expressive behavior as predictors of interpersonal
  consequences: A meta-analysis.
\newblock \emph{Psychological bulletin}, 111(2):256.

\bibitem[{Bartels(2006)}]{bartels}
Larry~M Bartels. 2006.
\newblock Priming and persuasion in presidential campaigns.
\newblock \emph{Capturing campaign effects}, 1:78--114.

\bibitem[{Borchers and Hundley(2018)}]{borchers2018rhetorical}
Timothy Borchers and Heather Hundley. 2018.
\newblock \emph{Rhetorical theory: An introduction}.
\newblock Waveland Press.

\bibitem[{Chaiken(1980)}]{chaiken1980heuristic}
Shelly Chaiken. 1980.
\newblock Heuristic versus systematic information processing and the use of
  source versus message cues in persuasion.
\newblock \emph{Journal of personality and social psychology}, 39(5):752.

\bibitem[{Cicero(1862)}]{cicero1862cicero}
Marcus~Tullius Cicero. 1862.
\newblock \emph{Cicero on Oratory and Orators}.
\newblock HG Bohn.

\bibitem[{Clark(2016)}]{clark_2016}
Roy~Peter Clark. 2016.
\newblock \emph{Writing Tools: 55 Essential Strategies for Every Writer}.
\newblock Little, Brown, and Co.

\bibitem[{Durmus and Cardie(2018)}]{durmus-cardie-2018-exploring}
Esin Durmus and Claire Cardie. 2018.
\newblock \href {https://doi.org/10.18653/v1/N18-1094} {Exploring the role of
  prior beliefs for argument persuasion}.
\newblock In \emph{Proceedings of the 2018 Conference of the North {A}merican
  Chapter of the Association for Computational Linguistics: Human Language
  Technologies, Volume 1 (Long Papers)}, pages 1035--1045, New Orleans,
  Louisiana. Association for Computational Linguistics.

\bibitem[{Durmus and Cardie(2019)}]{durmus-cardie-2019-corpus}
Esin Durmus and Claire Cardie. 2019.
\newblock \href {https://doi.org/10.18653/v1/P19-1057} {A corpus for modeling
  user and language effects in argumentation on online debating}.
\newblock In \emph{Proceedings of the 57th Annual Meeting of the Association
  for Computational Linguistics}, pages 602--607, Florence, Italy. Association
  for Computational Linguistics.

\bibitem[{Ebrahimi et~al.(2018)Ebrahimi, Rao, Lowd, and
  Dou}]{ebrahimi-etal-2018-hotflip}
Javid Ebrahimi, Anyi Rao, Daniel Lowd, and Dejing Dou. 2018.
\newblock \href {https://doi.org/10.18653/v1/P18-2006} {{H}ot{F}lip: White-box
  adversarial examples for text classification}.
\newblock In \emph{Proceedings of the 56th Annual Meeting of the Association
  for Computational Linguistics (Volume 2: Short Papers)}, pages 31--36,
  Melbourne, Australia. Association for Computational Linguistics.

\bibitem[{Finch(2000)}]{lyneyve2000}
Lyneyve Finch. 2000.
\newblock \href {https://doi.org/10.1177/0095327X0002600302} {Psychological
  propaganda: The war of ideas on ideas during the first half of the twentieth
  century}.
\newblock \emph{Armed Forces \& Society}, 26(3):367--386.

\bibitem[{Habernal and
  Gurevych(2016{\natexlab{a}})}]{habernal-gurevych-2016-makes}
Ivan Habernal and Iryna Gurevych. 2016{\natexlab{a}}.
\newblock \href {https://doi.org/10.18653/v1/D16-1129} {What makes a convincing
  argument? empirical analysis and detecting attributes of convincingness in
  web argumentation}.
\newblock In \emph{Proceedings of the 2016 Conference on Empirical Methods in
  Natural Language Processing}, pages 1214--1223, Austin, Texas. Association
  for Computational Linguistics.

\bibitem[{Habernal and
  Gurevych(2016{\natexlab{b}})}]{habernal-gurevych-2016-argument}
Ivan Habernal and Iryna Gurevych. 2016{\natexlab{b}}.
\newblock \href {https://doi.org/10.18653/v1/P16-1150} {Which argument is more
  convincing? analyzing and predicting convincingness of web arguments using
  bidirectional {LSTM}}.
\newblock In \emph{Proceedings of the 54th Annual Meeting of the Association
  for Computational Linguistics (Volume 1: Long Papers)}, pages 1589--1599,
  Berlin, Germany. Association for Computational Linguistics.

\bibitem[{Habernal and Gurevych(2017)}]{habernal-gurevych-2017-argumentation}
Ivan Habernal and Iryna Gurevych. 2017.
\newblock \href {https://doi.org/10.1162/COLI_a_00276} {Argumentation mining in
  user-generated web discourse}.
\newblock \emph{Computational Linguistics}, 43(1):125--179.

\bibitem[{Hovland et~al.(1953)Hovland, Janis, and
  Kelley}]{hovland1953communication}
Carl~Iver Hovland, Irving~Lester Janis, and Harold~H Kelley. 1953.
\newblock Communication and persuasion.

\bibitem[{Jang et~al.(2016)Jang, Gu, and Poole}]{jang2016categorical}
Eric Jang, Shixiang Gu, and Ben Poole. 2016.
\newblock Categorical reparameterization with gumbel-softmax.
\newblock \emph{arXiv preprint arXiv:1611.01144}.

\bibitem[{Kingma and Welling(2014)}]{kingma2013auto}
Diederik~P Kingma and Max Welling. 2014.
\newblock \href {https://arxiv.org/abs/1312.6114} {Auto-encoding variational
  bayes}.
\newblock In \emph{Proceedings of the 2nd International Conference on Learning
  Representations}, Banff, Canada. International Conference on Learning
  Representations.

\bibitem[{Macagno and Walton(2014)}]{macagno_walton_2014}
Fabrizio Macagno and Douglas Walton. 2014.
\newblock \emph{Emotive language in argumentation}.
\newblock Cambridge University Press.

\bibitem[{McHugh(2012)}]{mchugh2012interrater}
Mary~L McHugh. 2012.
\newblock Interrater reliability: the kappa statistic.
\newblock \emph{Biochemia medica: Biochemia medica}, 22(3):276--282.

\bibitem[{Mikolov et~al.(2013)Mikolov, Chen, Corrado, and
  Dean}]{mikolov2013efficient}
Tomas Mikolov, Kai Chen, Greg Corrado, and Jeffrey Dean. 2013.
\newblock Efficient estimation of word representations in vector space.
\newblock \emph{arXiv preprint arXiv:1301.3781}.

\bibitem[{Morio et~al.(2019)Morio, Egawa, and
  Fujita}]{morio-etal-2019-revealing}
Gaku Morio, Ryo Egawa, and Katsuhide Fujita. 2019.
\newblock \href {https://doi.org/10.18653/v1/D19-1653} {Revealing and
  predicting online persuasion strategy with elementary units}.
\newblock In \emph{Proceedings of the 2019 Conference on Empirical Methods in
  Natural Language Processing and the 9th International Joint Conference on
  Natural Language Processing (EMNLP-IJCNLP)}, pages 6274--6279, Hong Kong,
  China. Association for Computational Linguistics.

\bibitem[{Oraby et~al.(2015)Oraby, Reed, Compton, Riloff, Walker, and
  Whittaker}]{oraby-etal-2015-thats}
Shereen Oraby, Lena Reed, Ryan Compton, Ellen Riloff, Marilyn Walker, and Steve
  Whittaker. 2015.
\newblock \href {https://doi.org/10.3115/v1/W15-0515} {And that{'}s a fact:
  Distinguishing factual and emotional argumentation in online dialogue}.
\newblock In \emph{Proceedings of the 2nd Workshop on Argumentation Mining},
  pages 116--126, Denver, CO. Association for Computational Linguistics.

\bibitem[{Paszke et~al.(2019)Paszke, Gross, Massa, Lerer, Bradbury, Chanan,
  Killeen, Lin, Gimelshein, Antiga, Desmaison, Kopf, Yang, DeVito, Raison,
  Tejani, Chilamkurthy, Steiner, Fang, Bai, and Chintala}]{NEURIPS2019_9015}
Adam Paszke, Sam Gross, Francisco Massa, Adam Lerer, James Bradbury, Gregory
  Chanan, Trevor Killeen, Zeming Lin, Natalia Gimelshein, Luca Antiga, Alban
  Desmaison, Andreas Kopf, Edward Yang, Zachary DeVito, Martin Raison, Alykhan
  Tejani, Sasank Chilamkurthy, Benoit Steiner, Lu~Fang, Junjie Bai, and Soumith
  Chintala. 2019.
\newblock \href
  {http://papers.neurips.cc/paper/9015-pytorch-an-imperative-style-high-performance-deep-learning-library.pdf}
  {Pytorch: An imperative style, high-performance deep learning library}.
\newblock In H.~Wallach, H.~Larochelle, A.~Beygelzimer, F.~d\textquotesingle
  Alch\'{e}-Buc, E.~Fox, and R.~Garnett, editors, \emph{Advances in Neural
  Information Processing Systems 32}, pages 8024--8035. Curran Associates, Inc.

\bibitem[{Pedregosa et~al.(2011)Pedregosa, Varoquaux, Gramfort, Michel,
  Thirion, Grisel, Blondel, Prettenhofer, Weiss, Dubourg, Vanderplas, Passos,
  Cournapeau, Brucher, Perrot, and Duchesnay}]{scikit-learn}
F.~Pedregosa, G.~Varoquaux, A.~Gramfort, V.~Michel, B.~Thirion, O.~Grisel,
  M.~Blondel, P.~Prettenhofer, R.~Weiss, V.~Dubourg, J.~Vanderplas, A.~Passos,
  D.~Cournapeau, M.~Brucher, M.~Perrot, and E.~Duchesnay. 2011.
\newblock Scikit-learn: Machine learning in {P}ython.
\newblock \emph{Journal of Machine Learning Research}, 12:2825--2830.

\bibitem[{Popkin and Popkin(1994)}]{popkin1994reasoning}
Samuel~L Popkin and Samuel~L Popkin. 1994.
\newblock \emph{The reasoning voter: Communication and persuasion in
  presidential campaigns}.
\newblock University of Chicago Press.

\bibitem[{Schegloff and Sacks(1973)}]{schegloff1973opening}
Emanuel~A Schegloff and Harvey Sacks. 1973.
\newblock Opening up closings.
\newblock \emph{Semiotica}, 8(4):289--327.

\bibitem[{Srinivasan et~al.(2019)Srinivasan, Danescu-Niculescu-Mizil, Lee, and
  Tan}]{10.1145/3359265}
Kumar~Bhargav Srinivasan, Cristian Danescu-Niculescu-Mizil, Lillian Lee, and
  Chenhao Tan. 2019.
\newblock \href {https://doi.org/10.1145/3359265} {Content removal as a
  moderation strategy: Compliance and other outcomes in the changemyview
  community}.
\newblock \emph{Proc. ACM Hum.-Comput. Interact.}, 3(CSCW).

\bibitem[{Tan et~al.(2016)Tan, Niculae, Danescu-Niculescu-Mizil, and
  Lee}]{Tan_2016}
Chenhao Tan, Vlad Niculae, Cristian Danescu-Niculescu-Mizil, and Lillian Lee.
  2016.
\newblock \href {https://doi.org/10.1145/2872427.2883081} {Winning arguments}.
\newblock \emph{Proceedings of the 25th International Conference on World Wide
  Web - WWW ’16}.

\bibitem[{Wallace et~al.(2019)Wallace, Feng, Kandpal, Gardner, and
  Singh}]{wallace2019universal}
Eric Wallace, Shi Feng, Nikhil Kandpal, Matt Gardner, and Sameer Singh. 2019.
\newblock \href {https://doi.org/10.18653/v1/D19-1221} {Universal adversarial
  triggers for attacking and analyzing {NLP}}.
\newblock In \emph{Proceedings of the 2019 Conference on Empirical Methods in
  Natural Language Processing and the 9th International Joint Conference on
  Natural Language Processing (EMNLP-IJCNLP)}, pages 2153--2162, Hong Kong,
  China. Association for Computational Linguistics.

\bibitem[{Walton(1992)}]{walton_1992}
Douglas~N. Walton. 1992.
\newblock \emph{The place of emotion in argument}.
\newblock Pennsylvania State University Press.

\bibitem[{Wolf et~al.(2019)Wolf, Debut, Sanh, Chaumond, Delangue, Moi, Cistac,
  Rault, Louf, Funtowicz, and Brew}]{Wolf2019HuggingFacesTS}
Thomas Wolf, Lysandre Debut, Victor Sanh, Julien Chaumond, Clement Delangue,
  Anthony Moi, Pierric Cistac, Tim Rault, R'emi Louf, Morgan Funtowicz, and
  Jamie Brew. 2019.
\newblock Huggingface's transformers: State-of-the-art natural language
  processing.
\newblock \emph{ArXiv}, abs/1910.03771.

\bibitem[{Yang et~al.(2019)Yang, Chen, Yang, Jurafsky, and
  Hovy}]{yang-etal-2019-lets}
Diyi Yang, Jiaao Chen, Zichao Yang, Dan Jurafsky, and Eduard Hovy. 2019.
\newblock \href {https://doi.org/10.18653/v1/N19-1364} {Let{'}s make your
  request more persuasive: Modeling persuasive strategies via semi-supervised
  neural nets on crowdfunding platforms}.
\newblock In \emph{Proceedings of the 2019 Conference of the North {A}merican
  Chapter of the Association for Computational Linguistics: Human Language
  Technologies, Volume 1 (Long and Short Papers)}, pages 3620--3630,
  Minneapolis, Minnesota. Association for Computational Linguistics.

\bibitem[{Yang and Kraut(2017)}]{yang2017persuading}
Diyi Yang and Robert~E Kraut. 2017.
\newblock Persuading teammates to give: Systematic versus heuristic cues for
  soliciting loans.
\newblock \emph{Proceedings of the ACM on Human-Computer Interaction},
  1(CSCW):1--21.

\bibitem[{Yang et~al.(2017)Yang, Hu, Salakhutdinov, and
  Berg-Kirkpatrick}]{yang2017improved}
Zichao Yang, Zhiting Hu, Ruslan Salakhutdinov, and Taylor Berg-Kirkpatrick.
  2017.
\newblock \href {http://proceedings.mlr.press/v70/yang17d.html} {Improved
  variational autoencoders for text modeling using dilated convolutions}.
\newblock In \emph{Proceedings of the 34th International Conference on Machine
  Learning}, volume~70 of \emph{Proceedings of Machine Learning Research},
  pages 3881--3890, International Convention Centre, Sydney, Australia. PMLR.

\end{thebibliography}
\bibliographystyle{acl_natbib}

\newpage

\appendix

\section{Additional Model and Training Details}
For all models, hyperparameters were manually tuned using macro-averaged F-1 scores and early stopping as our selection criterion. Models were trained using an NVIDIA RTX 2080 Ti. For our optimizers, unless otherwise specified, we use AdamW with learning rate 1e-3, betas $(0.9, 0.999)$, eps 1e-08, and weight decay 0.01. We used PyTorch~\cite{NEURIPS2019_9015} and HuggingFace~\cite{Wolf2019HuggingFacesTS} for any deep learning work.
\label{appendix_hyper}

\begin{table}[t]
\centering
\begin{tabular}{@{}lr@{}}  %
\toprule
Model & Macro F-1 \\ 
\midrule
Random & .50  \\
Naive Bayes & .62  \\
\begin{tabular}[t]{@{}l@{}}BERT\end{tabular} & .64  \\
\textbf{VAE LSTM} & .61 \\
\bottomrule
\end{tabular}
\caption{Request label performance on Validation Set.}
\label{testperf}
\end{table}

\paragraph{VAE  + LSTM: }
We minimize the following objective function for our graphical model:

$$\begin{aligned}
\mathbb{E}_{\mathbf{l} \sim q(\mathbf{l} | \mathbf{x})}\left[\mathbb{E}_{\mathbf{z} \sim q(\mathbf{z} | \mathbf{x}, \mathbf{l})}[\log p(\mathbf{x} | \mathbf{z}, \mathbf{l})]\right.\\
-\mathrm{KL}[q(\mathbf{z} | \mathbf{x}, \mathbf{l}) \| p(\mathbf{z})]] \\
-\mathrm{KL}[q(\mathbf{l} | \mathbf{x}) \| p(\mathbf{l})]
\end{aligned}$$

We also use LSTMs for the inference $q(z|x,l)$, generative $p(x|l,z)$, and discriminative $q(l|x)$ networks for our VAE. We set the size of $z$ to be 64. $l$'s size is defined by the number of unique strategies in our dataset: 6. We use the reparameterization trick in \citet{kingma2013auto} to use backprop on $z$; and use Gumbel's softmax \cite{jang2016categorical} to model $l$ continuously. Finally, we use CBOW Word2Vec embeddings \cite{mikolov2013efficient} of size 128 to learn initial word embeddings. Our VAE was trained for 100 epochs; and our LSTM was trained for 50 epochs.

\begin{table*}[t]
\centering
\begin{tabular}{@{}rllrr@{}}  %
\toprule
Rank & Shorthand & Expansion & Avg. Attention & Success Rate \\%
\midrule
1 & Po Po EOS & Politeness, Politeness, EOS & 0.00 & 0.82 \\
2 & Re Po EOS & Reciprocity, Politeness, EOS & 0.01 & 0.76 \\
3 & Co Po EOS & Concreteness, Politeness, EOS & 0.02 & 0.71 \\
4 & Cr Po EOS & Credibility, Politeness, EOS & 0.01 & 0.70 \\
5 & SOS Re EOS & SOS, Reciprocity, EOS & 0.01 & 0.65 \\
6 & SOS Po EOS & SOS, Politeness, EOS & 0.00 & 0.65 \\
7 & SOS Im EOS & SOS, Impact, EOS & 0.03 & 0.51 \\
8 & Re Co EOS & Reciprocity, Concreteness, EOS & 0.04 & 0.51 \\
9 & SOS Co EOS & SOS, Concreteness, EOS & 0.02 & 0.50 \\
10 & SOS Im Re & SOS, Impact, Reciprocity & 0.04 & 0.50 \\
11 & Co Re EOS & Concreteness, Reciprocity, EOS & 0.05 & 0.50 \\
12 & SOS Co Po & SOS, Concreteness, Politeness & 0.04 & 0.50 \\
13 & SOS Co Re & SOS, Concreteness, Reciprocity & 0.04 & 0.49 \\
14 & SOS Im Co & SOS, Impact, Concreteness & 0.05 & 0.48 \\
15 & SOS Im Po & SOS, Impact, Politeness & 0.04 & 0.46 \\
16 & Re Co Po & Reciprocity, Concreteness, Politeness & 0.05 & 0.46 \\
17 & SOS Co Im & SOS, Concreteness, Impact & 0.07 & 0.43 \\
18 & Co Co EOS & Concreteness, Concreteness, EOS & 0.05 & 0.42 \\
19 & Re Co Co & Reciprocity, Concreteness, Concreteness & 0.08 & 0.41 \\
20 & Po Co Co & Politeness, Concreteness, Concreteness & 0.08 & 0.39 \\
21 & Co Co Re & Concreteness, Concreteness, Reciprocity & 0.08 & 0.39 \\
22 & Im Co Re & Impact, Concreteness, Reciprocity & 0.08 & 0.38 \\
23 & SOS Co Co & SOS, Concreteness, Concreteness & 0.07 & 0.38 \\
24 & SOS Co Cr & SOS, Concreteness, Credibility & 0.05 & 0.37 \\
25 & Co Im Re & Concreteness, Impact, Reciprocity & 0.07 & 0.36 \\
26 & Co Co Po & Concreteness, Concreteness, Politeness & 0.07 & 0.36 \\
27 & Im Co Co & Impact, Concreteness, Concreteness & 0.10 & 0.35 \\
28 & Co Co Cr & Concreteness, Concreteness, Credibility & 0.07 & 0.35 \\
29 & Co Im Co & Concreteness, Impact, Concreteness & 0.09 & 0.34 \\
30 & Co Co Co & Concreteness, Concreteness, Concreteness & 0.11 & 0.31 \\
31 & Co Co Im & Concreteness, Concreteness, Impact & 0.11 & 0.27 \\
\bottomrule
\end{tabular}
\caption{Extended List of Strategy Triple Attentions Ranked by Corresponding Success Rates. \textbf{SOS} indicates start of request; \textbf{EOS} indicates the end of a request.}
\label{strat_list}
\end{table*}

\paragraph{BERT Baseline:}
For our BERT Baseline, we finetune the small BERT Base Cased model, using the AdamW optimizer with learning rate of 2e-5 and Adams epsilon of 1e-8. Our BERT model is imported from HuggingFace's transformers repository, and was finetuned for 10 epochs.

\paragraph{Naive Bayes Baseline:}
We use the Multinomial Naive Bayes model, implemented with scikit-learn \cite{scikit-learn}, using default parameters.

\paragraph{Random Baseline:}
We use the dummy classifier with the random setting provided in Scikit-Learn \cite{scikit-learn}.

Validation performance across all classifiers can be seen in Table~\ref{testperf}.

\section{Persuasion Strategy Triplets}
A full ranked list of persuasion strategy triplets, along with average strategy attention and success rates can be found in Table \ref{strat_list}.

\end{document}